\documentclass[12pt]{article}

\newif\iffinal
\finaltrue

\newif\ifarxiv
\arxivtrue

\usepackage{sbc-template}

\usepackage{graphicx,url}

\usepackage[english]{babel}
\usepackage[T1]{fontenc} %
\usepackage[utf8]{inputenc} %
\usepackage{multicol}
\usepackage{multirow}
\usepackage{graphicx,url}
\usepackage{amsmath,amssymb,amsfonts}
\usepackage{algorithmic}
\usepackage{textcomp}
\usepackage{url}
\usepackage{enumerate}
\usepackage{multirow}
\usepackage{soul}
\usepackage{verbatim}
\usepackage[table,xcdraw,dvipsnames]{xcolor}
\usepackage[shortlabels]{enumitem}
\usepackage{scalefnt} %
\usepackage{comment} %
\usepackage{lscape}
\usepackage{fancyhdr}
\usepackage{sbgames}
\usepackage[super]{nth}
\usepackage{booktabs}
\usepackage[caption=false,farskip=0pt]{subfig}
\usepackage[colorlinks=true,allcolors=black]{hyperref}
\usepackage[capitalize,noabbrev]{cleveref}
\usepackage[acronym]{glossaries}
\usepackage{xspace}

\newacronymstyle{long-short-br}
{%
  \GlsUseAcrEntryDispStyle{long-short}%
}%
{%
  \GlsUseAcrStyleDefs{long-short}%
}
\setacronymstyle{long-short-br}

\usepackage{tikz}
\usepackage{transparent}

\ifarxiv
    \newcommand\publicationtext{%
        \scriptsize Accepted at SBGames 2026. The version of record is available in SBC OpenLib: \href{https://doi.org/10.5753/sbgames.2026.26163}{\textcolor{blue}{10.5753/sbgames.2026.26163}}.}
    \newcommand\publicationnotice{%
        \begin{tikzpicture}[remember picture,overlay]
            \node[anchor=south,yshift=24pt] at (current page.south) {%
                \fbox{%
                    \transparent{0.85}%
                    \parbox{\dimexpr0.835\textwidth-\fboxsep-\fboxrule\relax}{\publicationtext}%
                }%
            };
        \end{tikzpicture}%
    }
\fi

\emergencystretch=\maxdimen
\newcommand*{\RL}[2][]{\textcolor{Rhodamine}{[\textbf{\ifthenelse{\equal{#1}{}}{RL}{RL(#1)}}: #2]}}
\newcommand*{\RC}[2][]{\textcolor{Orange}{[\textbf{\ifthenelse{\equal{#1}{}}{RC}{RC(#1)}}: #2]}}

\trilha{Computing} %
\local{Goiânia/GO} %
\ano{2026} %
\edicao{XXV} %

\title{Computer Vision for MOBA Analytics: A Dataset and Baseline for Visibility Analysis in Dota~2}

\iffinal

    \author{Ricardo da Rocha Carvalho\inst{1}, Eloísa Oliveira\inst{1}, Luiz Bernardo Martins Kummer\inst{1},\\Emerson Cabrera Paraiso\inst{1}, Rayson Laroca\inst{1}
    }

    \address{Pontifícia Universidade Católica do Paraná (PUCPR), Curitiba, Brasil\\
        \tt\scriptsize \{ricardo.rcarvalho,oliveira.eloisa,luiz.kummer,paraiso,rayson\}@ppgia.pucpr.br\\
    }
\else
    \author{No Author Given\inst{1}}
    \address{No Institute Given}
\fi

\begin{document} 

\maketitle

\ifarxiv
    \pagestyle{empty}
    \thispagestyle{empty}
    \publicationnotice
\else
    \thispagestyle{plain}
\fi

\newacronym{api}{API}{Application Programming Interface}
\newacronym{dota2}{Dota~2}{Defense of the Ancients~2}
\newacronym{iou}{IoU}{Intersection over Union}
\newacronym{moba}{MOBA}{Multiplayer Online Battle Arena}

\newcommand{\dataset}{Dota2-Vis\xspace}
\newcommand{\supplementary}{\url{https://github.com/RicardoRCarvalho/dota2-vis/}}

\begin{abstract}
\textbf{Introduction:} Most \gls*{moba} analytics studies rely on structured data, which does not directly capture what each team could actually see during a match. 
\textbf{Objective:} This work introduces \dataset, a video-based dataset, and a baseline pipeline for visibility analysis in professional \acrshort*{dota2} matches. 
\textbf{Methodology:} The dataset comprises all 144 matches from \textit{The International~2025}, recorded from both team perspectives, totaling 288 Full HD~videos, together with 2,477 manually annotated minimap images. We evaluate multiple variants of a modern object detector for player-icon detection and use the best-performing model to estimate opponent-visible player presence over time. 
\textbf{Results:} YOLO11l~(large) achieved the best overall performance, reliably identifying player icons even in dense and visually cluttered minimap scenes. The resulting visibility curves reveal player, hero, role, and team-level patterns that complement conventional \gls*{moba} analytics, highlighting behavioral differences that are difficult to obtain from structured data alone. The \dataset dataset is publicly available at {\small \supplementary}.
\end{abstract}

     
\section{Introduction}
\glsresetall
\glsunset{dota2}

Game analytics has become an important tool for understanding player behavior, supporting decision-making, and extracting actionable knowledge from digital games~\cite{ElNasr2013GameAnalytics}. 
This is particularly relevant in esports, where matches generate large volumes of behavioral, spatial, and temporal information that can be used to characterize strategies, compare players, and support predictive models. 
Among esports genres, \glspl*{moba} are especially relevant for artificial intelligence research because they combine long-term planning, team coordination, real-time decision-making, and imperfect information~\cite{Katona2019TimeToDie,Costa2024AIinMOBA}.

Most prior work on \gls*{moba} analytics has relied primarily on structured data from drafts, event logs, replay files, and public \glspl*{api}, supporting tasks such as draft recommendation, match outcome prediction, team composition analysis, event prediction, and live win prediction~\cite{Summerville2016DraftAnalysis,Semenov2017PredictingGameOutcome,Hodge2021LivePrediction,Yang2022HoK,Yang2023PredictingEvents}. 
These sources remain central, but capture only part of match dynamics. 
Visual information can complement them by revealing what players can actually observe during a match. 
In \gls*{dota2}, for instance, player coordinates indicate where a hero is located, but not necessarily what the opposing team can see. 
Opponent visibility depends on fog of war, terrain elevation, trees, wards, unit vision ranges, temporary abilities, and other map-dependent factors, making it difficult to infer accurately from tabular data~alone.

Computer vision offers a complementary way to study these dynamics directly from gameplay footage, using the visual information presented to players during the match. 
This perspective is especially useful for analyzing visibility cues that depend on spatial context, transient game elements, and patch-dependent map changes. 
Recent studies have begun to explore visual information in \glspl*{moba}, including player camera analysis in \gls*{dota2}~\cite{Tot2021PlayerCam} and player trajectory extraction in \textit{League of Legends} videos~\cite{Kim2025LOL}. 
Nevertheless, to the best of our knowledge, no publicly available dataset currently provides the necessary annotations or structure to systematically study visibility-related information in professional \gls*{moba}~matches.

In this work, we present \dataset, a video-based dataset designed to support visibility analysis in professional \gls*{dota2} matches. 
It brings together dual-perspective recordings of all matches from \textit{The International~2025} and manually annotated minimap images collected from professional matches. 
The annotations include player icons, clones, and other relevant minimap elements, providing fine-grained data for training object detectors under visually dense \gls*{moba} conditions. 
We then explore this dataset to build a baseline pipeline that detects player icons on the minimap and estimates when each player is visible from the opposing team's perspective throughout a~match.

By aggregating detections over time, we obtain temporal descriptors of opponent-visible map presence for players, heroes, roles, and teams. 
These descriptors can reveal stylistic differences among players using the same hero and role, hero and role-dependent activation windows, and team-level visibility patterns associated with match outcomes.

The main contributions of this paper are threefold:
\begin{itemize}
    \item We introduce \dataset, a dataset composed of professional \gls*{dota2} match videos from both teams' perspectives and manually annotated minimap images for player/icon detection;
    \item We provide baseline object detection experiments for detecting player icons and related minimap elements in visually dense \gls*{moba} scenes;
    \item We provide initial visibility-based analyses that characterize opponent-visible map presence across players, roles, heroes, teams, match stages, and match outcomes. Although not exhaustive, these analyses illustrate how visibility descriptors extracted from video can be leveraged in practice for post-match review, scouting, opponent preparation, and data-driven strategic analysis.
\end{itemize}

The remainder of this paper is organized as follows.
\cref{sec:dota2} presents the main \gls*{dota2} characteristics relevant to this study.
\cref{sec:related_work} reviews related work on artificial intelligence, game analytics, and computer vision in \glspl*{moba}.
\cref{sec:dataset} describes the dataset, annotations, and detection baseline.
\cref{sec:results} presents the visibility-based analyses.
Finally, \cref{sec:conclusions} concludes the paper and outlines future~work.

\section{Dota 2}
\label{sec:dota2}

\acrfull*{dota2} is a \gls*{moba} in which two teams, Dire and Radiant, compete on a shared map. 
Each team has five players, and each player controls a unique hero with specific abilities, attributes, and strategic roles.
Competitive lineups are commonly organized into five roles: the hard carry, who receives resources to become a late-game damage threat; the mid laner, who gains early levels and helps control the pace of the early game; the offlaner, who contests space and initiates fights; and two supports, often referred to as position~4 and position~5, who provide utility, map control, and vision. 
The main objective is to destroy the opposing team's Ancient, a central structure located inside its base. 
Professional matches usually last between 30 and 40 minutes~\cite{Katona2019TimeToDie} and involve different phases, including laning, resource acquisition, objective control, team fights, and base~sieges.

A central characteristic of \gls*{dota2} is imperfect information. 
Teams only observe regions revealed by allied heroes, units, buildings, wards, and other temporary sources of vision, while the remaining areas are hidden by fog of war. 
Thus, visibility is a strategic resource: knowing where opponents are, inferring where they might be, and denying information to the opposing team directly affects movement, objective control, and team-fight preparation~\cite{Pedrassoli2020WARDS, Tot2021PlayerCam}.

\begin{figure}[!htb]
    \centering
    \captionsetup[subfigure]{labelformat=empty}

    \resizebox{0.99\linewidth}{!}{
    \subfloat[]{
        \includegraphics[height=18ex]{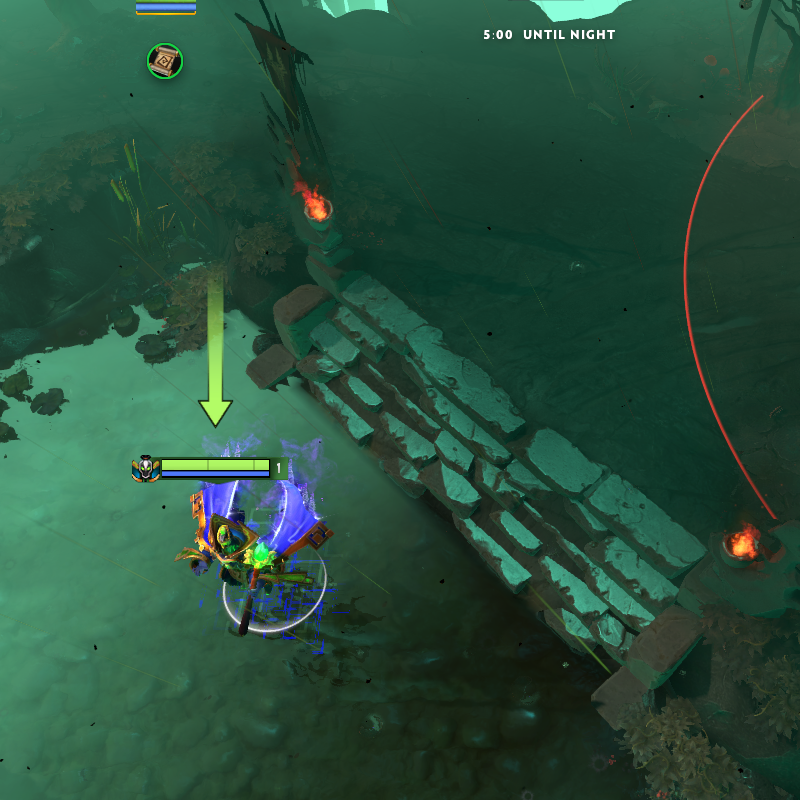}
        \includegraphics[height=18ex]{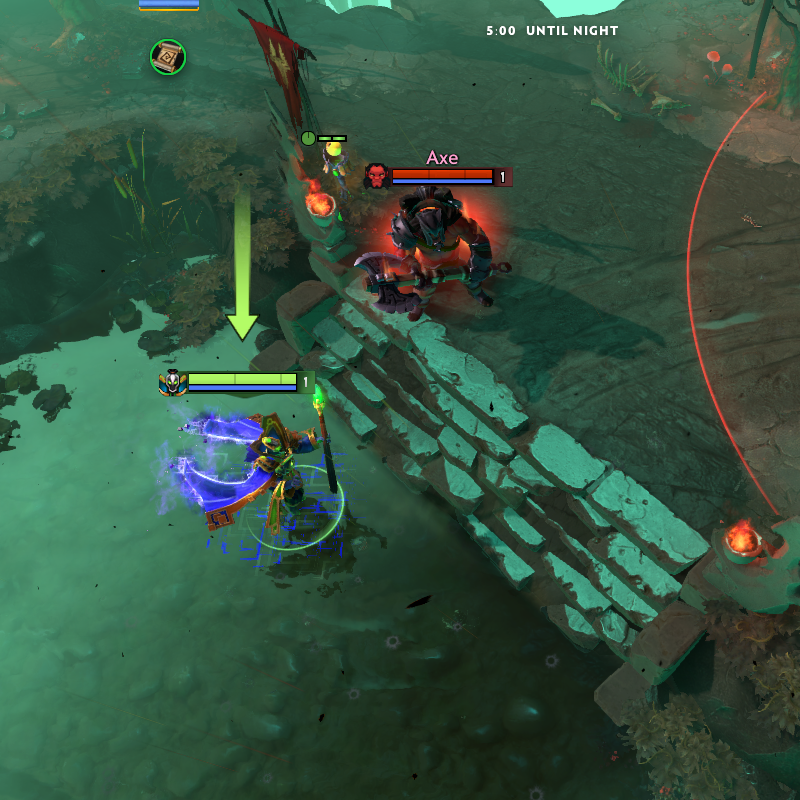}
    }\qquad
    \subfloat[]{
        \includegraphics[height=18ex]{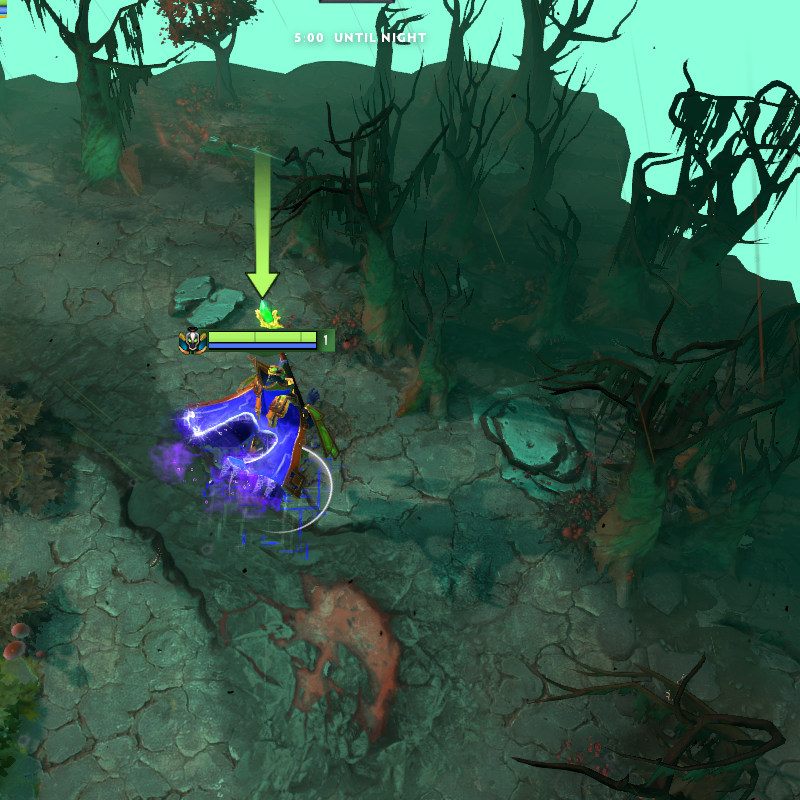}
        \includegraphics[height=18ex]{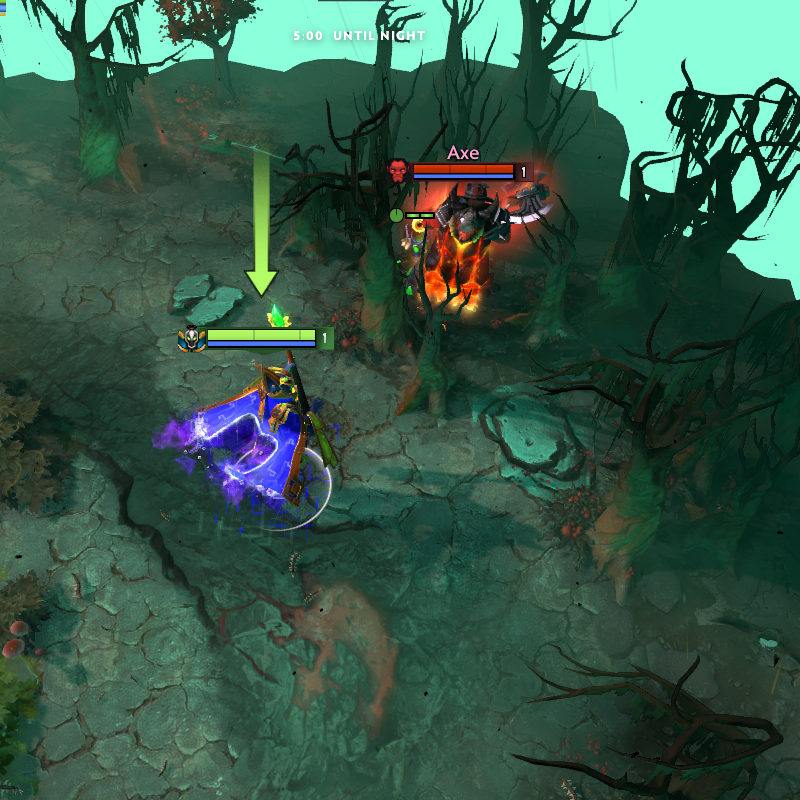}
    }
    }

    \vspace{-5.5mm}

    \caption{Representative examples illustrating why visibility in \gls*{dota2} cannot be inferred from player coordinates alone. Within each pair, the left image shows the information available from a player or team perspective, while the right image shows the complete game state, highlighting the effect of terrain elevation, trees, and local map geometry.}
    \label{fig:samples-vision}
\end{figure}

These mechanics make visibility a state-dependent signal rather than a simple function of hero location. 
As illustrated in \cref{fig:samples-vision}, nearby heroes are not necessarily mutually aware, since line of sight may be blocked or enabled by local map elements such as trees, terrain transitions such as stairs, and temporary vision~sources. 

\section{Related Work}
\label{sec:related_work}

Prior work on MOBA analytics has largely relied on structured match data, such as drafts, event logs, replay files, and public \glspl*{api}. 
In DotA 2, this line of research includes draft analysis and hero recommendation~\cite{Summerville2016DraftAnalysis}, outcome prediction from team composition~\cite{Semenov2017PredictingGameOutcome}, and live win prediction in professional matches~\cite{Hodge2021LivePrediction}. 
Similar predictive tasks have also been explored in other MOBA, including interpretable win prediction for \textit{Honor of Kings}~\cite{Yang2022HoK} and event prediction in MOBA matches~\cite{Yang2023PredictingEvents}. 
Together, these studies show the relevance of machine learning for esports analysis, while also illustrating the common dependence on information already available through logs, replays, or telemetry.

Other studies have investigated more specific aspects of match dynamics. 
\cite{Katona2019TimeToDie} and \cite{RINGER2023TimeToDie2}, for example, addressed in-game death prediction in DotA, while \cite{Pedrassoli2020WARDS} modeled the value of vision through warding information in \textit{League of Legends} and \gls*{dota2}. 
The latter is particularly related to our work, as it highlights the strategic importance of visibility in \glspl*{moba}. 
However, it estimates the value of vision from available game data, whereas we aim to recover visibility information directly from gameplay footage.

Computer vision remains less explored than structured-data approaches in \gls*{moba} research.
\cite{Tot2021PlayerCam} modeled player intent in \gls*{dota2} by combining hero and player-camera positions to predict team fights shortly before they occur, while \cite{Kim2025LOL} proposed an object-detection-based framework~(adapting YOLOv5) to extract player trajectories from \textit{League of Legends} videos.
Beyond \glspl*{moba}, \cite{Joo2025ViewingExperience} studied hierarchical scene detection and tracking in \textit{StarCraft} esports videos, aiming to improve automatic observing and spectator experience.
These studies suggest that visual information can complement traditional game data, especially when relevant events or states are unavailable or difficult to recover from structured~form.

Our work follows this direction, but focuses on a different aspect of \gls*{dota2}: player visibility.
By detecting player positions on the minimap from videos recorded from both teams' perspectives, we estimate when players are visible to opponents and analyze how this information evolves throughout professional matches.
To our knowledge, no publicly available \gls*{moba} dataset combines video data, dual-team perspectives, and minimap annotations in a way that supports this type of opponent-visibility analysis.
Thus, the proposed dataset and baseline experiments provide an initial step toward using computer vision to study visibility patterns that are difficult to obtain from tabular data~alone.

\section{The \dataset Dataset: Data Collection and Player Detection}
\label{sec:dataset}

We introduce \textit{\dataset}, a dataset designed to support visibility analysis in professional \gls*{dota2} matches. 
The dataset has two complementary components: (i)~full match videos recorded from both teams' perspectives, used for visibility analysis; and (ii)~manually annotated minimap images, used to train and evaluate the player detector. 
Both components, along with their annotations and predefined data splits, are publicly~available\footnote{\supplementary}.

To build the video component, we replayed professional matches using the official game engine and recorded them with OBS Studio, keeping the screen, game, and recording resolutions fixed at $1920 \times 1080$ pixels. 
As our goal is to analyze visibility at the highest level of play, we recorded all matches from \textit{The International~2025}, the main annual tournament in the professional \gls*{dota2} scene~\cite{TILiquipedia}. 
The tournament comprised 144 matches, each recorded twice, once from the Radiant perspective and once from the Dire perspective, resulting in 288 Full HD~videos.

To train the player detector, we collected 2,477 minimap images from professional matches that were not part of \textit{The International~2025}, preventing overlap between the training data and the matches analyzed later in this work~\cite{barz2020do}.
We focus on the minimap rather than the full gameplay screen because it provides a compact, standardized representation of what each team can observe at any given moment. 
This choice reduces the influence of camera movement, zoom, combat effects, and hero animations, while preserving the information required for our target task: estimating whether players are visible to the opposing~team.

The annotation protocol was designed to account for both regular players and game-specific visual elements that may interfere with visibility estimation. 
We initially considered 10 player classes, one for each player in the match. 
However, some heroes, items, and runes can create controllable \textit{clones}, i.e., additional units that may appear on the minimap similarly to the original hero. 
As clones may affect visibility analysis, we added one clone class for each player, resulting in 20 player-related classes. 
In addition, players can communicate with their teammates through \textit{pings}, which are temporary minimap markers used, for example, to indicate danger, movement, or attack intentions. 
As pings share the same color as the player who created them and may be visually confused with player icons, we added an \textit{Other} class to represent non-player elements that could affect detection. 
Thus, the final annotation protocol contains 21 classes. 
To avoid making the detector dependent on specific heroes, we configured the game to display players on the minimap using color and symbol icons instead of hero portraits. 
Notably, this makes the detector independent of the current hero pool and avoids the need for retraining whenever new heroes are added to the~game.

\begin{figure}[!htb]
    \centering

    \resizebox{0.99\linewidth}{!}{
        \includegraphics[width=0.24\linewidth]{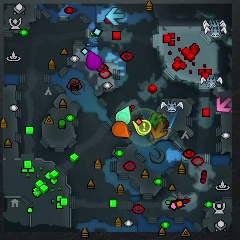}
        \includegraphics[width=0.24\linewidth]{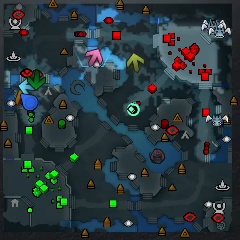}
        \includegraphics[width=0.24\linewidth]{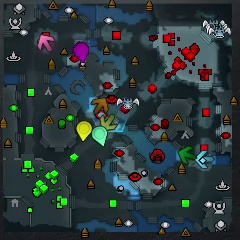}
        \includegraphics[width=0.24\linewidth]{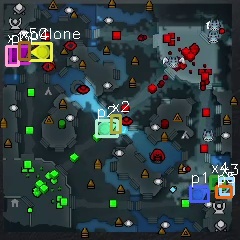}
    }

    \vspace{0.4mm}

    \resizebox{0.99\linewidth}{!}{
        \includegraphics[width=0.24\linewidth]{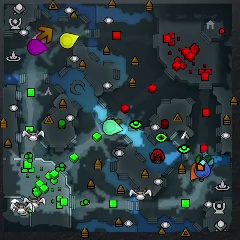}
        \includegraphics[width=0.24\linewidth]{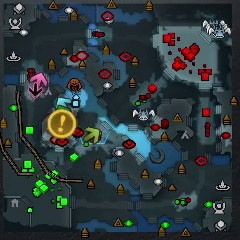}
        \includegraphics[width=0.24\linewidth]{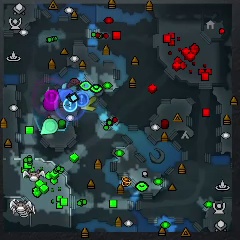}
        \includegraphics[width=0.24\linewidth]{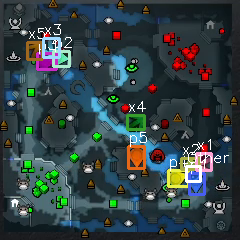}
    }
    \caption{Representative examples from the proposed \dataset dataset, illustrating the variability of minimap samples in terms of team side, game moment, object density, and visual clutter. The rightmost column shows manually labeled bounding boxes for~reference.}
    \label{fig:samples-dataset}
\end{figure}

All minimap images have resolution $240 \times 240$ pixels and were manually annotated with bounding boxes~(\textit{x}, \textit{y}, \textit{w}, \textit{h}).
This process is time-consuming because minimaps often contain many small, overlapping objects amid visual clutter and similar colors.
As shown in \cref{fig:samples-dataset}, the samples cover different match moments, ranging from sparse situations to dense scenes with several visible players and non-player elements, including challenging cases with clones, pings, and cluttered regions.
The rightmost column shows the corresponding bounding boxes, illustrating the level of detail required during annotation.
To improve reliability, the annotations were reviewed in multiple rounds using different data splits, with the remaining inconsistencies observed in preliminary experiments mostly restricted to difficult cases involving severe occlusion or visual~ambiguity.

The 2,477 images were split into training, validation, and test subsets, with the test set containing only frames from matches not used for training or validation.
This reduces visual overlap between training and evaluation data.
The final split contains 1,670 training, 407 validation, and 400 test images, following an approximately 70/15/15~distribution.

We adopted YOLOv11~\cite{yolov11} in the detection experiments, motivated by the strong performance of the YOLO family across diverse object detection tasks~\cite{terven2023comprehensive,Kim2025LOL,laroca2025improving}.
YOLOv11 provides five variants, from YOLO11n~(nano) to YOLO11x~(extra-large), which differ in size, complexity, and speed-accuracy trade-off.
Considering that the most suitable variant depends on the target application, we evaluate all five YOLO11 variants in this~work.

The hyperparameters were selected based on preliminary experiments conducted on the validation set. 
The final training configuration used 200 epochs with early stopping patience of 15, image size of $480\times480$ pixels, a maximum of 40 detections per image, and dropout of 0.1. 
We also applied data augmentation, including small changes in saturation and brightness, vertical flips, mosaic augmentation, scale variation, erasing, and small geometric transformations, such as rotation, shear, and perspective~changes.

For evaluation, we report precision, recall, F-score, and mAP$_{50:95}$.
A detection is considered correct when its predicted class matches the ground truth and the overlap between the predicted and annotated bounding boxes satisfies \gls*{iou}~$> 0.5$, following the standard protocol adopted in object detection benchmarks~\cite{everingham2010pascalvoc}.
We place particular emphasis on F-score and mAP$_{50:95}$.
The F-score is the harmonic mean of precision and recall, capturing the balance between false detections and missed players, both of which directly affect visibility estimation.
In turn, mAP$_{50:95}$ evaluates detection quality across multiple \gls*{iou} thresholds, providing a stricter assessment of localization performance~\cite{lin2014microsoft}.

As shown in \cref{tab:yolov11_results}, the YOLO11l~(large) variant achieved the best overall results, obtaining the highest F-score and mAP$_{50:95}$ among the evaluated variants. 
Although YOLO11x is substantially larger than YOLO11l~(194.9 BFLOPs vs.\ 86.9 BFLOPs), it did not improve performance in this setting. 
One possible explanation is that the task involves small and visually simple objects, while the available training data may not be sufficient for the additional model capacity to translate into better generalization. 
Thus, we selected YOLO11l as the model used in the visibility analyses reported in this~work.

\begin{table}[!htb]
    \centering
    \caption{Results obtained by the different YOLOv11 variants. The \textit{large} variant provided the best trade-off between computational cost and accuracy.}
    \label{tab:yolov11_results}

    \vspace{-1.75mm}

    \resizebox{0.675\linewidth}{!}{
    \begin{tabular}{lcccc}
        \toprule
        \textbf{Model} & \textbf{Precision} & \textbf{Recall} & \textbf{F-score} & \textbf{mAP$_{50:95}$} \\
        \midrule
        YOLO11n~(nano) & 0.964 & 0.881 & 0.921 & 0.661 \\
        YOLO11s~(small) & 0.965 & 0.896 & 0.929 & 0.713 \\
        YOLO11m~(medium) & 0.963 & 0.903 & 0.932 & 0.721 \\
        \textbf{YOLO11l~(large)} & 0.974 & 0.906 & \textbf{0.939} & \textbf{0.729} \\
        YOLO11x~(extra-large) & 0.965 & 0.911 & 0.937 & 0.720 \\
        \bottomrule
    \end{tabular}
    }
\end{table}

\begin{figure}[!htb]
    \centering

    \includegraphics[width=0.24\linewidth]{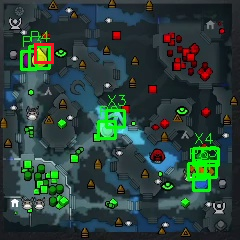}
    \includegraphics[width=0.24\linewidth]{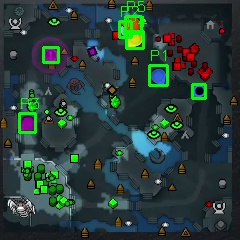}
    \includegraphics[width=0.24\linewidth]{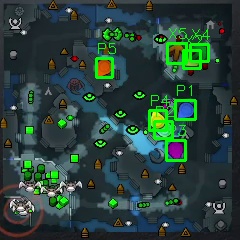}
    \includegraphics[width=0.24\linewidth]{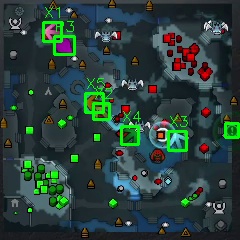}

    \caption{Representative detections produced by YOLO11l~(large). Green bounding boxes denote correct detections, whereas the red bounding box in the leftmost image highlights a missed detection of a heavily occluded player icon. Overall, the model detects players reliably, with failures occurring mainly under severe occlusion.}
    \label{fig:samples-yolo-results}
\end{figure}

\cref{fig:samples-yolo-results} shows representative detections from the selected model. 
Overall, the detector identifies players reliably, even in dense minimap scenes and visually cluttered situations. 
Most errors occur in rare cases of severe occlusion or strong overlap between small objects, as illustrated by the missed detection highlighted in red in the~figure.

Importantly, as an additional target-domain sanity check, we manually annotated 400 held-out minimap frames from \textit{The International~2025} videos, sampling two to three random frames per game.
The detector achieved performance consistent with \cref{tab:yolov11_results}, with an F-score of~0.940 and an mAP$_{50:95}$ of~0.729.

All models were implemented using the Ultralytics framework~\cite{yolov11}. 
The experiments were run on a computer equipped with an NVIDIA RTX 4090 GPU with 24GB of memory.

\section{Visibility-Based Game Analytics}
\label{sec:results}

The previous section established the feasibility of detecting player icons on the \gls*{dota2} minimap. 
We now shift the focus from detector performance to the type of game analytics enabled by the proposed pipeline. 
Rather than using visual data only as an auxiliary representation, our goal is to recover a match signal that is difficult to obtain from structured data alone: which players were visible to the opposing team over time. 
This information complements traditional \gls*{moba} analytics based on drafts, events, public \glspl*{api}, or match statistics~\cite{ElNasr2013GameAnalytics,Costa2024AIinMOBA}, and is especially relevant because visibility is directly related to map control, pressure, and uncertainty in decision-making~\cite{Pedrassoli2020WARDS}.

For each match, we process the videos recorded from both team perspectives. 
A player is considered visible when their minimap icon is detected in the video corresponding to the opponent's perspective.
We aggregate these detections in five-minute intervals and compute the visibility percentage as:
\begin{equation}
    \mathrm{Vis}(\mathcal{E}, I) =
    100 \cdot
    \frac{1}{|\mathcal{F}_I|\,|\mathcal{E}|}
    \sum_{f \in \mathcal{F}_I}
    \sum_{e \in \mathcal{E}}
    \mathbf{1}(e \in \mathcal{D}_f),
\end{equation}
where $\mathcal{F}_I$ is the set of frames in interval $I$, $\mathcal{E}$ is the entity or group of entities being analyzed, and $\mathcal{D}_f$ is the set of detections in frame $f$. 
Depending on the choice of $\mathcal{E}$, this metric can represent the visibility of a player, a hero, a role, or an entire team. 
In the following analyses, we report curves up to the 45-minute mark to reduce the influence of late-game intervals represented by fewer matches.

\begin{figure}[!htb]
    \centering
    \captionsetup[subfigure]{captionskip=0pt}

    \subfloat[\label{fig:graphs_a}]{
        \includegraphics[width=0.49\linewidth]{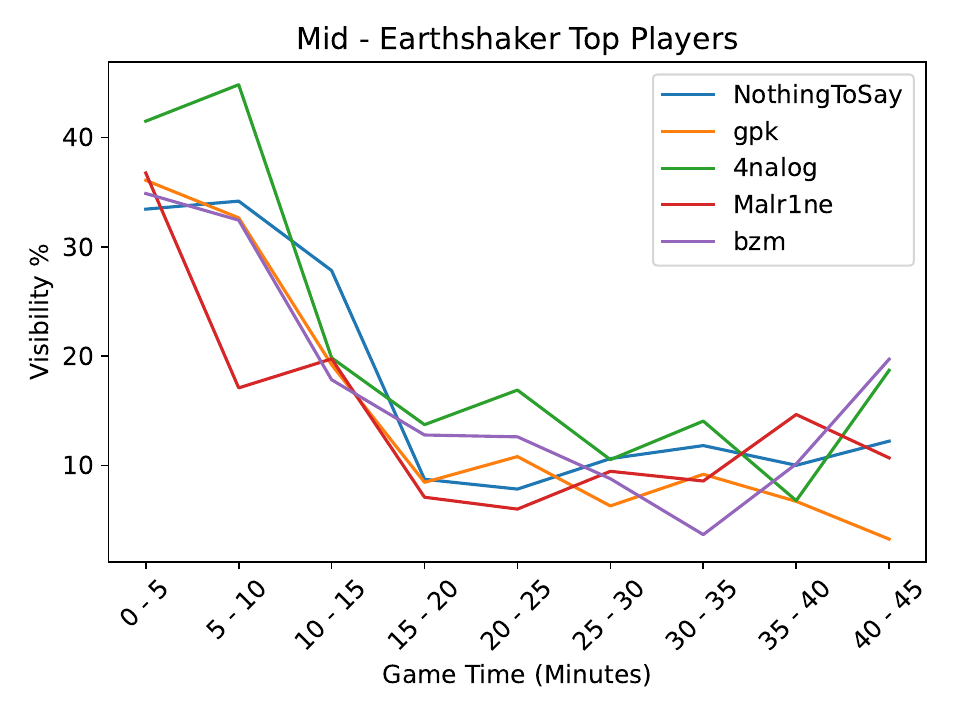}
    }
    \subfloat[\label{fig:graphs_b}]{
        \includegraphics[width=0.49\linewidth]{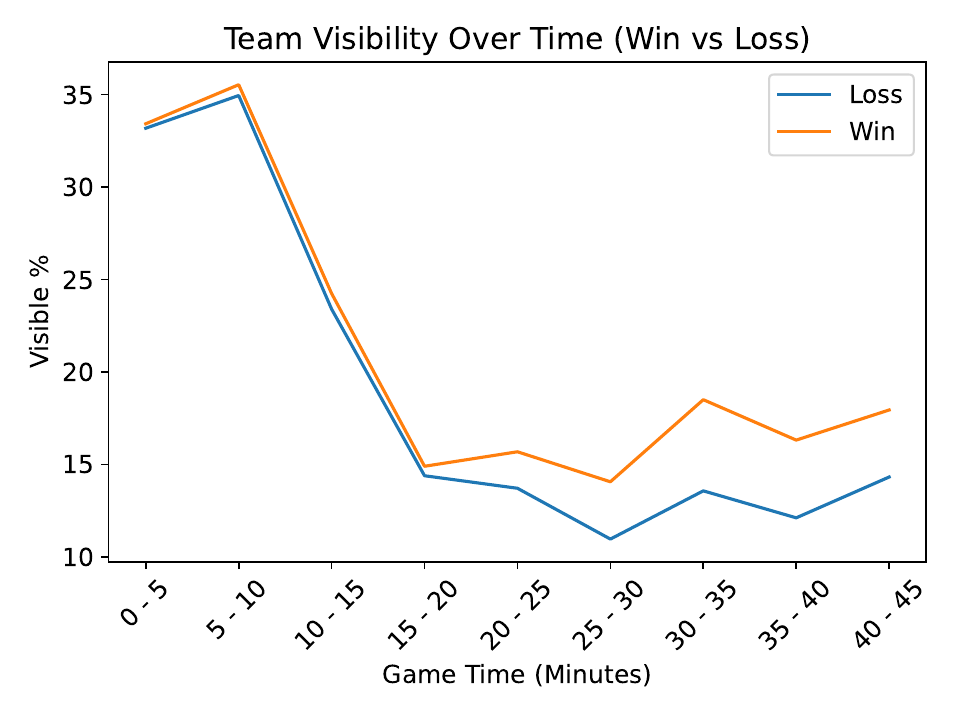}
    }

    \vspace{-0.5mm}
    
    \caption{Examples of visibility-based analyses enabled by the proposed pipeline. 
    (a)~Player-level visibility curves for the mid-lane \textit{Earthshaker }players with the largest number of matches in the tournament. 
    (b)~Team-level visibility curves grouped by match outcome. 
    Curves are computed in five-minute intervals from videos of \textit{The International 2025}.}
    \label{fig:graphs}
\end{figure}

\cref{fig:graphs_a} illustrates a player-level analysis in which the hero and role are fixed. 
We selected \textit{Earthshaker}, the most picked mid-lane hero in the tournament, and compared the five players with the largest number of matches using this hero, with each player appearing in at least three matches. 
This setting reduces part of the variability caused by hero choice and allows the visibility curves to highlight differences in player behavior. 
For example, \textit{4nalog}, one of the analyzed players, presents higher early-game visibility than the others, suggesting a more active or pressure-oriented use of the hero during the first minutes. 
In contrast, player \textit{Malr1ne} remains less visible in the same period, which may indicate a more conservative approach, safer resource acquisition, or a strategy based on keeping the opponent uncertain about his position. 
These curves should not be interpreted as direct measures of player quality. 
Instead, they provide a quantitative way to describe stylistic differences usually discussed qualitatively by analysts, coaches, and~players. 

At the team level, \cref{fig:graphs_b} shows that winning and losing teams have similar visibility during the early game, especially during the laning stage. 
After approximately 20 minutes, however, the curves start to separate, with winning teams becoming more visible on the map. 
A plausible interpretation is that, once a team gains resources and map control, it can occupy more contested areas, pressure objectives, and force the opposing team to play from safer or hidden positions. 
This behavior is consistent with the snowballing dynamics commonly observed in \glspl*{moba}, in which an initial advantage is progressively converted into more control over space and resources~\cite{li2017visual}. 
From a modeling perspective, this suggests that opponent-visible map presence could be used as an additional feature for live win prediction or event prediction, complementing structured indicators such as gold, experience, kills, objectives, and draft information~\cite{Semenov2017PredictingGameOutcome,Hodge2021LivePrediction,Yang2022HoK,Yang2023PredictingEvents}.

\begin{figure}[!htb]
    \centering
    \captionsetup[subfigure]{captionskip=0pt}

    \subfloat[\label{fig:graphs_heroes_a}]{
        \includegraphics[width=0.49\linewidth]{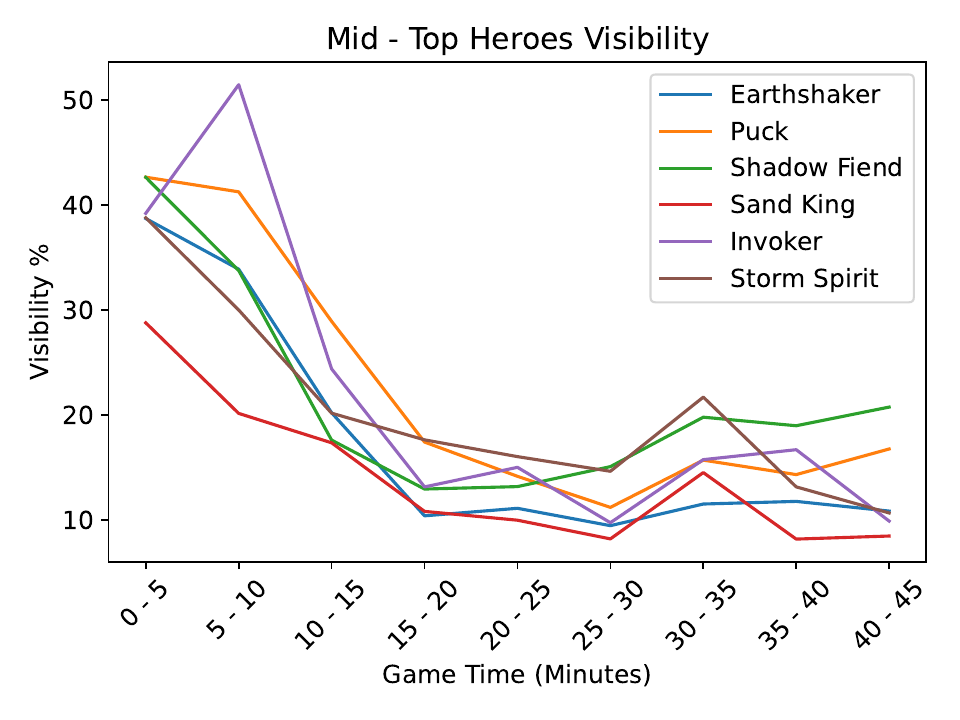}
    }
    \subfloat[\label{fig:graphs_heroes_b}]{
        \includegraphics[width=0.49\linewidth]{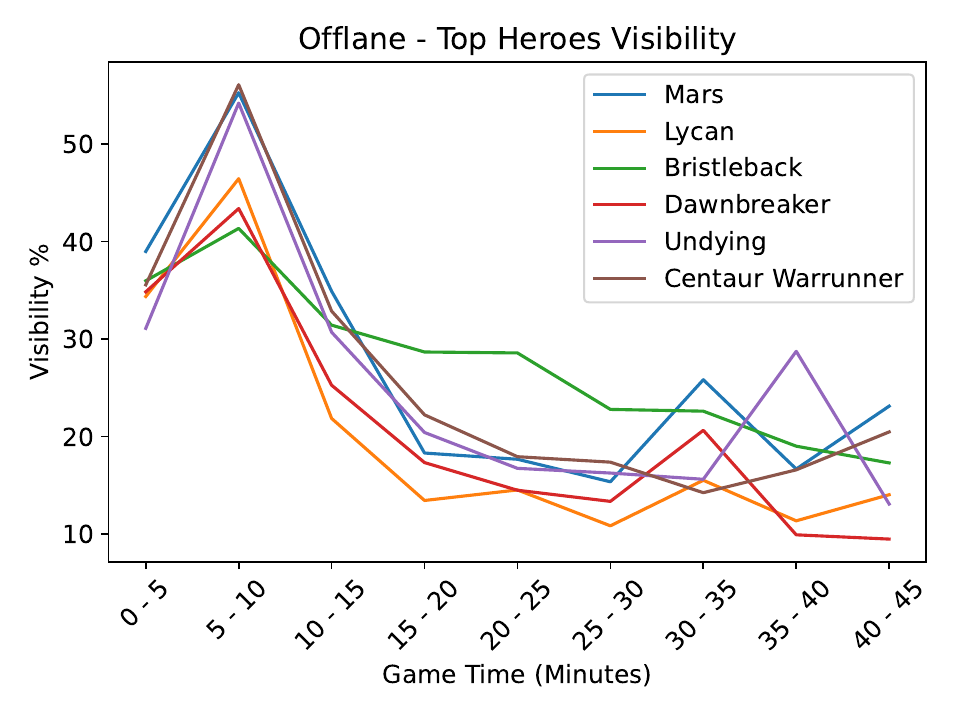}
    }

    \vspace{-0.5mm}
    
    \caption{Hero-level visibility curves for the six most picked heroes in two roles at \textit{The International 2025}. 
    (a)~Mid lane. 
    (b)~Offlane. 
    The curves illustrate how the proposed visual signal can reveal hero-specific activation windows and role-dependent map presence patterns.}
    \label{fig:graphs_heroes}
\end{figure}

The same representation can also be used to compare heroes and roles. 
\cref{fig:graphs_heroes} shows visibility curves for the six most picked heroes in the mid lane and offlane roles. 
In the mid lane~(\cref{fig:graphs_heroes_a}), heroes exhibit different activation patterns: some are more visible in the early game, while others become more present only after acquiring levels, items, or mobility. 
For instance, the curve of \textit{Storm Spirit} suggests a later increase in map activity, which is consistent with a hero that often requires resources before becoming a constant threat across the map. 
In the offlane~(\cref{fig:graphs_heroes_b}), visibility tends to reflect the role's responsibility of contesting space, initiating fights, and disrupting the enemy carry. 
Heroes such as \textit{Mars} and \textit{Bristleback} illustrate different timings of this pressure, with visibility changing as their relative strength evolves throughout the match. 
These observations should not be interpreted as patch-invariant conclusions about specific heroes.
Rather, they show that the proposed visual signal can transform strategic concepts such as activity, pressure, and hiddenness into measurable temporal patterns.

Beyond aggregate analysis, the recovered visibility signal can support practical forms of performance review and training. 
For players, it can be used to compare map presence against reference profiles from high-level players using the same hero and role, helping identify moments of unusual exposure, excessive passivity, or different activity timings. 
For teams, it can guide post-match review by highlighting intervals in which visible map presence did not translate into objectives, pressure, or resource denial, as well as moments in which low visibility may indicate missed opportunities to contest space. 
In opponent preparation, player-hero visibility profiles may also reveal recurring timing patterns, such as when a mid player tends to rotate, when an offlaner starts occupying dangerous areas, or when a team begins showing itself before major objectives. 
These uses should be understood as diagnostic support rather than automatic coaching prescriptions: visibility alone does not explain intent, but it can guide replay review and provide objective evidence for discussions about map control, risk, and timing~\cite{Novak2019PerformanceAnalysisEsports,Bialecki2024EsportsTraining}.

These examples highlight why the proposed \dataset dataset goes beyond a minimap detection benchmark. 
By extracting visibility directly from video, it provides behavioral descriptors that are difficult to recover from structured data alone. 
Player coordinates, for instance, do not necessarily indicate what the opposing team could actually see. 
As discussed in \cref{sec:dota2} and illustrated in \cref{fig:samples-vision}, visibility depends on occlusions and map geometry, including trees that may block vision and elevation changes that affect line of sight.
However, some trees can be destroyed during a match, and map layouts may change across game patches.
Hence, even a hard-coded visibility model would require constant updates and could still diverge from what was actually displayed to players during the match.
Our video-based approach therefore provides a complementary path for \gls*{moba} analytics, connecting recent efforts on visual understanding of esports footage~\cite{Tot2021PlayerCam,Kim2025LOL,Joo2025ViewingExperience} with strategic analysis in professional \gls*{dota2}.

\section{Conclusions}
\label{sec:conclusions}

This paper presented \dataset, a dataset for studying visibility in professional \gls*{dota2} matches through computer vision. 
The dataset comprises dual-perspective gameplay videos from \textit{The International~2025} and manually annotated minimap images, enabling the training and evaluation of methods that extract visibility-related information from gameplay footage. 
To our knowledge, \dataset is the first dataset to support this type of opponent-visibility analysis in professional \gls*{dota2} matches from~video.

We also introduced a baseline pipeline for player-icon detection on the minimap and for estimating when each player is visible to the opposing team. 
This pipeline converts a visual, context-dependent game mechanic into quantitative temporal descriptors. 
The results show that this formulation is feasible and provides a useful starting point for visibility-centered analysis, with YOLO11l~(large) achieving the best overall trade-off among the evaluated detectors. 
Initial analyses of the extracted visibility curves reveal patterns that are difficult to obtain from structured data alone, including differences among players using the same hero and role, hero- and role-dependent activation windows, and changes in visibility trends between winning and losing teams after the early game. 
These preliminary findings suggest that opponent-visible map presence may provide a useful signal for studying activity, pressure, risk, and map control in \gls*{dota2}.

The analyses reported in this work should be interpreted as an initial demonstration of what \dataset enables, rather than as a complete characterization of professional \gls*{dota2} visibility patterns. 
Given the scope of this work, we focus on representative cases that illustrate how the proposed visual signal can support analyses of players, heroes, roles, teams, match stages, and outcomes. 
Broader analyses across different matchups and contexts remain an important direction for future work. 
Moreover, as the dataset is centered on a single tournament, some findings may reflect the specific teams, patch, metagame, and broadcast conditions of \textit{The International~2025}.

Future work will expand the dataset to additional tournaments and patches, combine visibility with structured data from public \glspl*{api}, and incorporate temporal tracking and uncertainty estimation. 
These extensions would support more detailed analyses of team-level visibility profiles, objective-contest scenarios, smoke plays, ganks, warding patterns, and match-stage differences. 
More broadly, the same video-based formulation could be adapted to other \glspl*{moba}, such as \textit{League of Legends} and \textit{Honor of Kings}, further extending the use of computer vision for visibility-centered game analytics.

\section*{Acknowledgments}

This study was financed in part by the \textit{Coordenação de Aperfeiçoamento de Pessoal de Nível Superior -- Brasil} (CAPES), Finance Code~001.

\bibliographystyle{sbc}
\bibliography{bibtex}

\end{document}